\documentclass[twocolumn]{article}
\usepackage{natbib}
\usepackage{amsmath}
\usepackage{amsthm}
\usepackage{amsfonts}
\usepackage{amssymb}
\usepackage{url}
\usepackage{color}
\usepackage{subfigure}
\usepackage{nicefrac}
\usepackage{graphicx}
\usepackage[left=2cm,right=2cm,top=1.5cm,bottom=2cm,includefoot,includehead]{geometry}

\usepackage{verbatim}

\usepackage{hyperref}

\title{Quantifying Morphological Computation based on an Information Decomposition of the Sensorimotor Loop}
\author{Keyan Ghazi-Zahedi$^{1}$, Johannes Rauh$^{2}$
\mbox{}\\
$^1$Max Planck Institute for Mathematics in the Scienes, Inselstrasse 22, 04103
Leipzig, Germany\\
$^2$
Leibniz Universit{\"a}t Hannover, Welfengarten 1, 30167 Hannover, Germany \\
zahedi@mis.mpg.de, rauh@math.uni-hannover.de}

\begin{document}
\maketitle
\begin{abstract}
The question how an agent is affected by its embodiment has attracted growing attention in recent years. A new field of artificial intelligence has emerged, which is based on the idea that intelligence cannot be understood without taking into account embodiment. We believe that a formal approach to quantifying the embodiment's effect on the agent's behaviour is beneficial to the fields of artificial life and artificial intelligence. The contribution of an agent's body and environment to its behaviour is also known as morphological computation. Therefore, in this work, we propose a quantification of morphological computation, which is based on an information decomposition of the sensorimotor loop into shared, unique and synergistic information. In numerical simulation based on a formal representation of the sensorimotor loop, we show that the unique information of the body and environment is a good measure for morphological computation. The results are compared to our previously derived quantification of morphological computation.
\end{abstract}

\section{Introduction}
Morphological computation is discussed in various contexts, such as DNA
computing and self-assembly \cite[see][for an
overview]{2007International-Conference-on-Morphological,%
Hauser2012Introduction-to-the-Special-Issue}. In this publication, we are
interested in quantifying morphological computation of embodied agents which are embedded in the
sensorimotor loop. Morphological computation, in this context, is described as the trade-off between
morphology and control~\citep{Pfeifer1999Understanding-intelligence}, which
means that a well-chosen morphology, if exploited, substantially reduces the
amount of required control \citep{MontufarAG}. Hereby, a morphology refers to
the agent's body, explicitly including all its physiological and physical
properties (shape, sensors, actuators, friction, mass distribution,
etc.)~\citep{Pfeifer2002Embodied-artificial-intelligence}. The consensus is that
morphological computation is the contribution of the morphology and environment to the behaviour, that
cannot be assigned to a nervous system or a controller. There are several 
examples from biology, which demonstrate how the behaviour of an agent relies on
the interaction of the body and environment. A nice example is given by
\citet[][see p.~188]{Wootton1992Functional-Morphology-of-Insect}, who describes
how ``active muscular forces cannot entirely control the wing shape in flight.
They can only interact dynamically with the aerodynamic and inertial forces that
the wings experience and with the wing's own elasticity; the instantaneous
results of these interactions are essentially determined by the architecture of
the wing itself [\ldots]''

One of the most cited example from the field of embodied artificial intelligence
is the Passive Dynamic Walker by \citet{McGeer1990Passive-dynamic-walking}. In
this example, a two-legged walking machine preforms a naturally appealing
walking behaviour, as a result of a well-chosen morphology and environment,
without any need of control. There is simply no computation available and the
walking behaviour is the result of the gravity, the slope of the ground and the
specifics of the mechanical construction (weight and length of the body parts,
deviation of the joints, etc.). If any parameter of the mechanics (morphology)
or the slope (environment) is changed, the walking behaviour will not persist. 
In this context, we understand the exploitation of the body's and environment's
physical properties as the embodiments effect on a behaviour.

Theoretical work on describing morphological computation in the context of
embodied artificial intelligence has been conducted by
\citep{Hauser2011Towards-a-theoretical-foundation,%
  Fuchslin2012Morphological-Computation-and-Morphological}. In this publication,
we study an information-theoretic approach to quantifying morphological
computation which is based on an information decomposition of the sensorimotor
loop. This work is based on two of our previous publications in which we have
investigated different quantifications of morphological computation
\citep{Zahedi2013aQuantifying} and derived a general decomposition of a mutual
information of three random variables into unique, shared, and synergistic
information \citep{BROJA13:Quantifying_unique_information}. In our previous work
\citep{Zahedi2013aQuantifying}, we derived two concepts which both match the
general intuition about morphological computation, but showed different results.
In this publication, we will apply the information decomposition of
\citet{BROJA13:Quantifying_unique_information} to the setting of
\citet{Zahedi2013aQuantifying} with the goal to unify the two previously derived
concepts.

The paper is organised in the following way. The next section discusses the
sensorimotor loop and its representation as a causal graph. The third section
describes the bivariate information decomposition
from~\cite{BROJA13:Quantifying_unique_information}. Based on the information
decomposition, the fourth section introduces the unique information as
a measure for morphological
computation in the sensorimotor loop. The fifth section presents numerical
results, which are then discussed in the final section. An appendix explains how
we computed our measure of morphological computation.

\section{Sensorimotor Loop}
\label{sec:sml}
Our information theoretic decomposition of the mutual information requires a
formal representation of the sensorimotor loop, which we will introduce in
this section. In our understanding, a cognitive system consists of a brain or
controller, which sends signals to the system's actuators, thereby affecting the
system's environment. We prefer the notion of the system's \emph{Umwelt}
\citep{Uexkuell1957A-Stroll-Through,Clark1996Being-There:-Putting,%
Zahedi2010Higher-coordination-with}, which is the part of the system's
environment that can be affected by the system, and which itself affects the
system. The state of the actuators and the \emph{Umwelt} are not directly
accessible to the cognitive system, but the loop is closed as information about
the \emph{Umwelt} and the body is provided to the controller through the
sensors. In addition to this general concept of the sensorimotor loop, which is
widely used in the embodied artificial intelligence community \citep[see
e.g.][]{Pfeifer2007Self-Organization-Embodiment-and} we introduce the notion of
\emph{world} and by that we mean the system's morphology and the system's
\emph{Umwelt}. We can now distinguish between the intrinsic and extrinsic
perspective in this context. The world is everything that is extrinsic from the
perspective of the cognitive system, whereas the controller, sensor and actuator
signals are intrinsic to the system. This is analogous to the agent-environment
distinction in the context of reinforcement learning
\citep{Sutton1998aReinforcement}, in which the environment is understood as
everything that cannot be controlled arbitrarily by the agent.

\begin{figure}[h]
  \begin{center}
    \includegraphics[width=0.9\columnwidth]{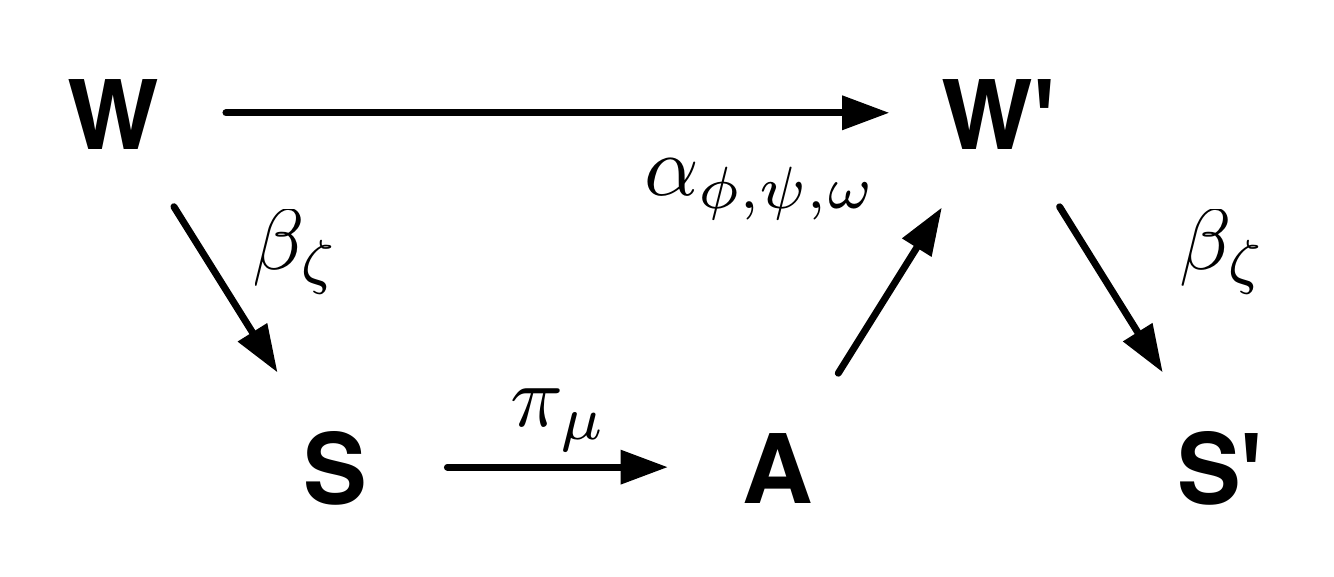}
  \end{center}
  \caption{A formal model of the sensorimotor loop.}\label{fig:binary_model} % Binary model
\end{figure}

The distinction between intrinsic and extrinsic is also captured in the
representation of the sensorimotor loop as a causal or Bayesian graph (see
Fig.~\ref{fig:binary_model}). For simplicity, we only discuss the sensorimotor
loop for reactive systems. This is plausible, because behaviours which exploit
the embodiment are usually better described as reactive and not as deliberative.
The most prominent examples are locomotion behaviours, e.g.~human walking,
swimming, flying, etc., which are all well-modelled as reactive behaviours.

The random variables $S$, $A$, and $W$ refer to sensor, actuator, and world
state, and the directed edges reflect causal dependencies between the random
variables \citep[see][]{Klyubin2004Organization-of-the-information-flow,%
Ay2008Information-Flows-in,Zahedi2010Higher-coordination-with}. Everything
that is extrinsic is captured in the variable $W$, whereas $S$ and $A$ are
intrinsic to the agent. The random variables $S$ and $A$ are not to be mistaken
with the sensors and actuators. The variable $S$ is the output of the sensors,
which is available to the controller or brain, the action $A$ is the input that
the actuators take. Consider an artificial robotic system as an example. Then
the sensor state $S$ could be the pixel matrix delivered by some camera sensor
and the action $A$ could be a numerical value that is taken by some motor
controller to be converted in currents to drive a motor.

Throughout this work, we use capital letter ($X$, $Y$, \ldots) to denote random
variables, non-capital letter ($x$, $y$, \ldots) to denote a specific value that
a random variable can take, and calligraphic letters ($\mathcal{X}$,
$\mathcal{Y}$, \ldots) to denote the alphabet for the random variables. This
means that $x_t$ is the specific value that the random variable $X$ can take a
time $t\in\mathbb{N}$, and it is from the set $x_t\in\mathcal{X}$. Greek letters
refer to generative kernels, i.e.~kernels which describe an actual underlying
mechanism or a causal relation between two random variables. 

We abbreviate the random variables for better comprehension in the remainder of
this work, as the information decomposition (see next sections) considers
random variables of consecutive time
indices. Therefore, we use the following notation. Random variables without any
time index refer to time $t$ and hyphened variables to time $t+1$.
The two variables $W,W'$ refer to $W_t$ and $W_{t+1}$.

Formally, the sensorimotor loop is given by the probability distribution $p(w)$
and the kernels $\alpha(w'|w,a)$, $\beta(s|w)$, and $\pi(a|a)$. To analyse
the quality of our derived quantification, it is best to evaluate them in a
fully controllable setting. For this purpose, we chose the same parameterisable
binary model of the sensorimotor loop that was used in our previous publication
\citep{Zahedi2013aQuantifying}. It allows to control the causal dependencies of $S$,
$A$, and $W$ individually, and thereby, enables an evaluation of the information
decomposition in the sensorimotor loop and compare it with our previous
results. The model is shown in Figure~\ref{fig:binary_model} and given by the
following set of equations:
\begingroup
\allowdisplaybreaks
\begin{align}
   \alpha_{\phi,\psi,\omega}(w'|w,a) & = \frac{e^{\phi w'w + \psi w'a +\omega w'wa}}%
  {\sum_{w''\in\Omega} e^{\phi w''w + \psi w''a+w''wa}}\label{eq:alpha}\\
  \beta_\zeta(s|w) & = \frac{e^{\zeta sw}}{\sum_{s''\in\Omega}e^{\zeta s''w}}\label{eq:beta}\\
  \pi_\mu(a|s)  & = \frac{e^{\mu as}}{\sum_{a'\in\Omega}e^{\mu a's}}\label{eq:pi}\\
  p_\tau(w) & = \frac{e^{\tau w}}{\sum_{w''\in\Omega}e^{\tau w''}}\label{eq:p},
\end{align}
\endgroup
where $a,w,s,w'\in\Omega=\{\pm1\}$ and $\phi,\psi,\omega,\zeta,\mu,\tau \geq 0$.
As in \citep{Zahedi2013aQuantifying}, the following two assumptions
are made without loss of generality. First, it is assumed that all world states
$w\in\Omega$ occur with
equal probability, i.e. $p(w=1) = p(w=-1) = \nicefrac12$. Second, we assume
a deterministic sensor, i.e.~$\zeta\gg1\Rightarrow p(s|w) = \delta_{sw}$, which
means that the sensor is a copy of the world state. The first assumption does
not violate the generality, because it only assures that the world state itself
does not already encode some structure, which is propagated through the
sensorimotor loop. The second assumption does not violate the generality of the
model, because in a reactive system as in
Figure~\ref{fig:binary_model}, the sensor state $S$ and $A$ can be reduced to
a common state, with a new generative kernel $\gamma(a|w) = \pi(a|s) \circ
\beta(s|w)$. Hence, keeping one of the two kernels deterministic and varying the
other in the experiments below, does not reduce the validity of
this model. This leaves four open parameters $\psi,\phi,\omega$, and $\mu$, against
which the morphological computation measure is validated.

% The next section introduces the information decomposition that underlies our measures of morphological computation.
%  before the experiments
% are presented and discussed.

\section{Information Decomposition}

Next, we introduce the information decomposition that underlies our measure of
morphological computation.  We first explain this information decomposition in a
general information theoretic setting and later explain how we use it in the
sensorimotor loop.

Consider three random variables $X,Y,Z$.  Suppose that a system wants to predict
the value of the random variable~$X$, but it can only access the information
in~$Y$ or~$Z$.  How is the information that $Y$ and~$Z$ carry about~$X$
distributed over~$Y$ and~$Z$?  In general, there may be \emph{redundant} or
\emph{shared} information (information contained both $Y$ and~$Z$), but there
may also be \emph{unique} information (information contained in only one of~$Y$
or~$Z$).  Finally, there is also the possibility of \emph{synergystic} or
\emph{complementary} information, i.e.~information that is only available when
$Y$ and~$Z$ are taken together.  The classical example for synergy is the XOR
function: If $Y$ and $Z$ are binary random variables and if~$X=Y
\operatorname{XOR} Z$, then neither $Y$ nor $Z$ contain any information
about~$X$ (in fact, $X$ is independent of~$Y$ and $X$ is independent of~$Z$),
but when $Y$ and~$Z$ are taken together, they completely determine~$X$ (in
particular, $X$ is not independent from the pair $(X,Y)$).

The total information that $(X,Y)$ contains about~$X$ can be quantified by the
mutual information $I(X:(Y,Z))$. However, there is no canonical way to separate
these different kinds of informations.  Mathematically, one would like to have
four functions $SI(X:Y;Z)$ (``shared information''), $UI(X:Y\setminus Z)$
(``unique information of~$Y$''), $UI(X:Z\setminus Y)$ (``unique information
of~$Z$''), $CI(X:Y;Z)$ (``complementary information'') that satisfy
\begin{multline}
  \label{eq:MI-decomposition}
  I(X:(Y,Z)) = SI(X:Y;Z) + UI(X:Y\setminus Z)
  \\  + UI(X:Z\setminus Y) + CI(X:Y;Z).
\end{multline}
From the interpretation it is also natural to require
\begin{equation}
  \label{eq:MI-decomposition-2}
  \begin{split}
    MI(X:Y) &= SI(X:Y;Z) + UI(X:Y\setminus Z), \\
    MI(X:Z) &= SI(X:Y;Z) + UI(X:Z\setminus Y).
  \end{split}
\end{equation}
A set of three functions $SI$, $UI$, and $CI$ that
satisfy~\eqref{eq:MI-decomposition} and~\eqref{eq:MI-decomposition-2} is called
a \emph{bivariate information decomposition}
by~\cite{BROJA13:Quantifying_unique_information}.  It follows from the defining
equations and the chain rule of mutual information that an information
decomposition always satisfies
\begin{equation}
  \label{eq:conditional-decomposition}
  MI(X:Y|Z) = UI(X:Y\setminus Z) + CI(X:Y;Z).
\end{equation}

Equations~\eqref{eq:MI-decomposition} and~\eqref{eq:MI-decomposition-2} do not
specify the functions $SI$, $UI$, and~$CI$.  Several different candidates have
been proposed so far, for example
by~\cite{WilliamsBeer:Nonneg_Decomposition_of_Multiinformation}
and~\cite{HarderSalgePolani13:Bivariate_redundant_information}.  We will use the
decomposition of~\cite{BROJA13:Quantifying_unique_information} that is defined
as follows:

Let $\Delta$  % =\Delta(\Xcal\times\Ycal\times\Zcal)
be the set of all possible joint distributions of $X$, $Y$, and~$Z$.
Fix an element $P\in\Delta$ (the ``true'' joint distribution of $X$, $Y$, and~$Z$).
Define
\begin{multline*}
  \Delta_{P} = \Big\{ Q\in\Delta : \\
  \shoveright{ Q(X=x,Y=y)=P(X=x,Y=y) }\\
  \shoveright{ \text{ and }Q(X=x,Z=z)=P(X=x,Z=z)} \\
  \text{ for all }x\in\mathcal{X},y\in\mathcal{Y},z\in\mathcal{Z} \Big\}
\end{multline*}
as the set of all joint distributions which have the same marginal distributions on the pairs $(X,Y)$ and $(X,Z)$.
Then
\begin{align*}
  UI(X:Y\setminus Z)
  & = \min_{Q\in\Delta_{P}} I_{Q}(X:Y|Z), \\
  SI(X:Y;Z) &= \max_{Q\in\Delta_{P}} CoI_{Q}(X;Y;Z), \\
  CI(X:Y;Z) &= I(X:(Y,Z)) - \min_{Q\in\Delta_{P}}I_{Q}(X:(Y,Z)),
\end{align*}
where $CoI$ denotes the co-information. %
Here, a subscript $Q$ in an information quantity means that the quantity is
computed with respect to~$Q$ as the joint distribution.

One idea behind these functions is the following: Suppose that the joint
distribution $P$ of $X$, $Y$, and $Z$ is not known, but that just the marginal
distributions of the pairs $(X,Y)$ and~$(X,Z)$ are known.  This information is
sufficient to characterize the set~$\Delta_{P}$, but we do not know which
element of~$\Delta_{P}$ is the true joint distribution.  One can argue that the
$UI$ and $SI$ should be constant on~$\Delta_{P}$; that is, shared information
and unique information should depend only on the interaction of~$X$ and~$Y$ and
the interaction on $X$ and~$Z$, but not on the way in which the three variables
interact.

The second property that characterizes the information decomposition is that the set $\Delta_{P}$ contains a
distribution~$Q$ such that $CI_{Q}(X:Y;Z)=0$.  In other words, when only the marginal distributions of the pairs $(X,Y)$
and~$(X,Z)$ are known, then we cannot know whether there is synergy or not.
See~\citep{BROJA13:Quantifying_unique_information} for a more detailed justification and a proof how these properties,
determine the functions $UI$, $SI$, and~$CI$.

In~\cite{BROJA13:Quantifying_unique_information}, the formulas for~$UI$, $CI$,
and~$SI$ are derived from considerations about decision problems in which the
objective is to predict the outcome of~$X$.  Here, we want to apply the
information decomposition in another setting: We will set $X=W'$, $Y=W$,
and~$Z=A$.  In our setting, $W$ and~$A$ not only have information about~$W'$,
but they actually \emph{control}~$W'$.  However, % from an abstract point of view,
the situation is similar: In the sensorimotor loop, we also expect to find
aspects of redundant, unique, and complementary influence of~$W$ and~$A$
on~$W'$.  Formally, since everything is defined probabilistically, we can still use the
same functions~$UI$, $CI$, and~$SI$.  We believe that the arguments behind the definition of $UI$, $CI$ and~$SI$ remain
valid in the setting of the sensorimotor loop where we need it.
First, it is still plausible that unique and redundant contributions should only depend on the marginal distributions of
the pairs $(W,W')$ and~$(A,W')$.  Second, in order to decide whether $W$ and $A$ act synergistically, it does not
suffice to know only these marginal distributions.
Therefore, we believe that the functions $UI$,
$CI$, and $SI$ have a meaningful interpretation.  In particular, we hope to be
able to use the information decomposition in order to measure morphological
computation.  This view is supported by our simulations below, which indicate that the functions $UI$, $CI$ and $SI$ do
indeed lead to a reasonable decomposition of $I(W':(A,W))$ and that the unique information $UI(W:W'\setminus A)$ is a
reasonable measure of morphological computation, at least in our simple model of the sensorimotor loop.

The parameters of our model of the sensorimotor loop (Eqs~\eqref{eq:alpha}
to~\eqref{eq:p}) can also be interpreted in terms of an information
decomposition.  Intuitively, $\phi$ corresponds to the unique influence of $W$
on~$W'$, $\psi$ corresponds to the unique influence of $A$ on~$W'$, and $\omega$
corresponds to the complementary influence.  However, the role of the additional
parameters~$\zeta,\mu,\tau$ is not so clear, and it is not so easy to find a
correspondence of redundant information.  The information decomposition has the
advantage, that its definition does not depend on a parametrization. Observe
that if the ``synergistic parameter'' $\omega=0$ vanishes, then it does not
necessarily follow that~$CI(W':A;W)=0$ (see Fig.~\ref{fig:mc}). However, we do
expect the complementary information to be small in this case.

\section{Morphological computation}
Morphological computation was described as the contribution of the embodiment
to a behaviour. In our previous work, we
derived two concepts to quantify morphological computation, which are both based
on the world dynamics kernel $\alpha(w'|w,a)$.

The first concept assumes that
the current action $A$ has no influence on the next world state $W'$, in which
case the kernel $\alpha(w'|w,a)$ reduces to $\hat{\alpha}(w'|w)$. If this is the
case, we would say that the systems shows maximal morphological computation, as
the behaviour is completely determined by the world. To measure the amount of
morphological computation present in a recorded behaviour, we calculated how
much the data differed from the assumption by calculating the weighted Kullback-Leibler
divergence $\sum_{w,a}p(w,a)D_{KL}(\alpha(w'|w,a)\|\hat{\alpha}(w'|w))$, which is the
conditional mutual information $I(W':A|W)$. Because this quantity is zero if we
have maximal morphological computation, we inverted and normalised in the
following way: $1-I(W':A|W)/\log_2|W|$.

The second concept started with the complementary assumption that the current
world state $W$ had no influence on the next world state $W'$, i.e., that the
world dynamics kernel is given by $\tilde{\alpha}(w'|a)$. Morphological
computation was then quantified as the error from the assumption, given by the
weighted Kullback-Leibler divergence $\sum_{w,a}p(w,a)D_{KL}(\alpha(w'|w,a)\|\tilde{\alpha}(w'|a))$,
which equals the conditional mutual information $I(W':W|A)$.

Both concepts were analysed and quantifications were derived, which didn't
require knowledge about the world, but could be calculated from intrinsically
available information only. At that time, we could not determine which of the
two concepts would capture morphological computation best, although both
concepts and their intrinsic adaptations lead to different results in a
specific configuration ($\psi=\phi\approx 0$).

Our intention in this publication is to answer this question. For this purpose,
we follow a different approach to quantify morphological computation, by
starting with the mutual information of $I(W':(W,A))$ and decompose it into
the shared, unique and synergistic information, as described in the previous section.
Rewriting the
Equation~\eqref{eq:MI-decomposition}, by replacing $X,Y,Z$ by $W',W,A$, we obtain
the following information decomposition:
\begin{align}
  % I(W';W,A) & = \underbrace{SI(W':W;A)}_{\text{shared inf.}} +
  % \underbrace{UI(W':W\setminus A)}_{\text{unique inf.\ } W', W}
  % \nonumber\\
            % & + \underbrace{UI(W':A\setminus W)}_{\text{unique inf.\ } W', A}
            % + \underbrace{SY(W':W;A)}_{\text{synergistic inf.}}
  I(W':(W,A)) & = SI(W':W;A) + UI(W':W\setminus A) \nonumber\\
            & + UI(W':A\setminus W) + SY(W':W;A)
\end{align}
As show in Equation~\eqref{eq:conditional-decomposition}, our previous concept
two, the conditional mutual information $I(W':W|A)$ is given by the sum of the
unique information $UI(W':W\setminus A)$ and the synergistic information
$CI(W':W;A)$:
\begin{align}
  \label{eq:condI-as-sum}
  I(W':W|A) = UI(W':W\setminus A) + CI(W':W;A).
\end{align}
The examples we have discussed in the introduction (insect wing and
Passive Dynamic Walker) suggest to use the unique information
$UI(W':W\setminus A)$ to quantify morphological computation, because it captures
the information that the current and next world state $W,W'$ share uniquely.
The next section presents numerical simulations to investigate how the
conditional mutual information $I(W':W|A)$ and the unique information
$UI(W':W\setminus W)$ compare with respect to quantifying morphological
computation. 
\section{Experiments}
\begin{figure}[t]
  \begin{center}
    \includegraphics[width=0.95\columnwidth]{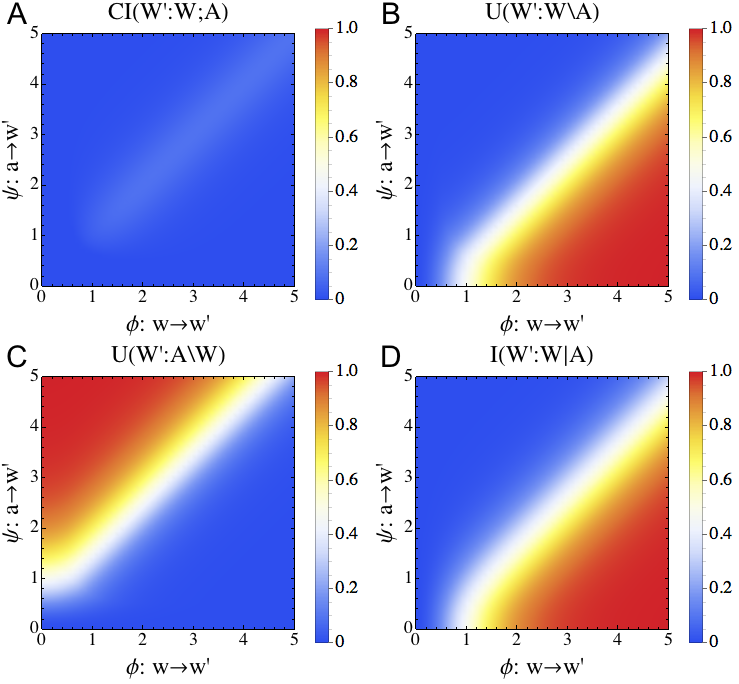}
  \end{center}
  \caption{Information decomposition for $\mu=0.0, \omega=0.0$}\label{fig:mc}
\end{figure}

\begin{figure}[t]
  \begin{center}
    \includegraphics[width=0.95\columnwidth]{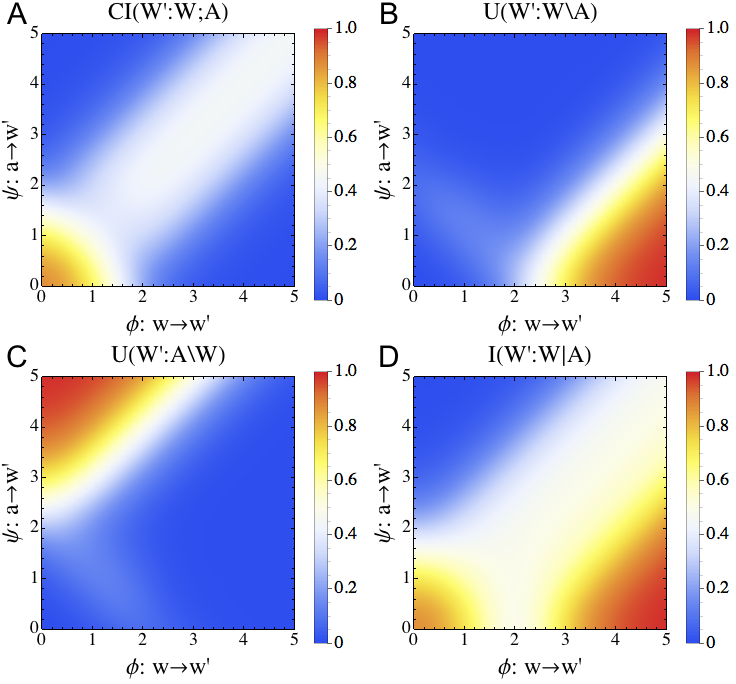}
  \end{center}
  \caption{Information decomposition for $\mu=0.0, \omega=2.0$}\label{fig:mc_2}
\end{figure}

The experiments in this section are conducted on the parameterised model of the
sensorimotor loop that was introduced in the second section (see
Fig.~\ref{fig:binary_model} and Eq.~\eqref{eq:alpha} to Eq.~\eqref{eq:p}). As stated
earlier, we set $\tau=0$, which means that the world state $W$ is drawn with
equal probability ($p(w=-1) = p(w=1) = \nicefrac12$), and $\zeta$ such that the
sensor state $S$ is a copy of the world state $W$. This leaves four parameters
for variation, namely the three world dynamics kernel parameters $\phi, \psi,
\omega$ and the policy parameter $\mu$. We decided
to plot the information theoretic quantities only for $\mu=0$ (see
Fig.~\ref{fig:mc} and Fig.~\ref{fig:mc_2}), i.e.,~for the case, in which the
action $A$ is chosen independently of the current sensor value $S$ and with
equal probability. This allows us to investigate the effect of the action $A$ on
the next world state $W'$, without any influence of $W$ on $A$. We also know
from previous experiments \citep[see][]{Zahedi2013aQuantifying}, that the
conditional mutual information $I(W':W|A)$ drops to zero for increasing
$\mu$. As the conditional mutual information is the sum of the unique and
synergistic information, we know that both quantities will also decrease with
increasing $\mu$. If $A$ is deterministically dependent on $W$, it also follows
that the unique information $UI(W':A\setminus W)$ is zero, because $A$ and $W$
are interchangeable. The only quantity that will be larger than zero is the
shared information, which, by definition, is not of interest in the context of
this work.

We decided to plot the information decomposition for varying $\phi$ (parameter
of unique influence of $W$ on $W'$) and $\psi$ (parameter of unique influence of
$A$ on $W'$) for two different values of $\omega$ (parameter of synergistic
influence of $W,A$ on $W'$, see Eq.~\eqref{eq:alpha}).
Figure~\ref{fig:mc} shows the results for
$\omega=0$, while Figure~\ref{fig:mc_2} shows the results for $\omega=2$. We
will first discuss the results for $\omega=0.0$, as they are best comparable
with our previous results from \citep{Zahedi2013aQuantifying}.
\par
\paragraph{Vanishing synergistic parameter ($\omega=0$):}
Figure~\ref{fig:mc}{\sf A} shows that synergistic information
$CI(W':W;A)$ is small and only present if $\psi \approx \phi$ (diagonal of
the image).  This is in agreement with our intuition that $\omega$ is the synergistic parameter.
The unique information of the action $A$ and the next world state
$W'$, denoted by $UI(W':A\setminus W)$, is shown in Figure~\ref{fig:mc}{\sf B}.
The plot reveals that the unique information of the current action $A$ and the
next world state $W'$ is only present, whenever $\psi>\phi$, and it is large,
whenever $\psi$ is significantly larger than $\phi$. Figure~\ref{fig:mc}{\sf C}
shows analogous results for the unique information 
$UI(W':W\setminus A)$. In this case, the unique information is negligible,
whenever $\phi \lesssim \psi$ and grows whenever $\phi$ is significantly larger
then $\psi$. These two plots show that the definition of the unique information,
as proposed by \citet{BROJA13:Quantifying_unique_information}, is able to
extract the unique influence in a setting in which two random variables % $Y,Z$
actually control a third random variable. % $X$. The fourth plot (see
Fig.~\ref{fig:mc}{\sf D} shows the conditional mutual information $I(W':W|A)$,
which was the second concept of quantifying morphological computation in our
previous work \citep{Zahedi2013aQuantifying}. As stated earlier, the conditional
mutual information is given by the sum of the unique and synergistic information
(Eq.~\eqref{eq:condI-as-sum}). Hence, there is almost no difference between
Figure~\ref{fig:mc}{\sf B} and Figure~\ref{fig:mc}{\sf D}, except on the
diagonal, where the unique information
$UI(W':W\setminus A)$ drops faster to zero.

\begin{figure}[t]
  \begin{center}
    \includegraphics[width=0.45\columnwidth]{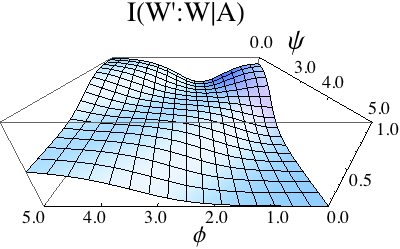}\hspace*{0.025\columnwidth}
    \includegraphics[width=0.45\columnwidth]{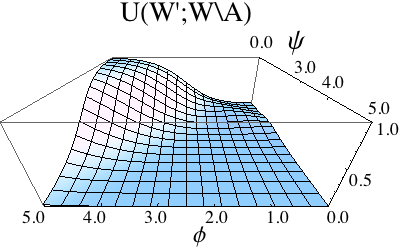}
  \end{center}
  \caption{Difference between $I(W':W|A)$ and $UI(W':W\setminus A)$ for
    $\omega=2$.}\label{fig:mi_vs_ui}
\end{figure}

\paragraph{Positive synergistic parameter ($\omega=2$):}
To study the difference between the unique information $UI(W':W\setminus
A)$ and the conditional mutual information $I(W':W|A)$, and hence, compare the
new quantification with our former concept, we conducted the same experiments
with a value of~$\omega=2$ (see Figures~\ref{fig:mc_2} and~\ref{fig:mi_vs_ui}).
Figures~\ref{fig:mc_2}{\sf A}-{\sf C} demonstrate
how the information decomposition can distinguish between the synergistic
information and the unique informations, which is exactly what we need %, if we want
to quantify morphological computation. The unique information
$UI(W':W\setminus A)$ captures only the information that the current world state
$W$ and the next world state $W'$ share, and therefore, captures the common
understanding of morphological computation in the context of embodied artificial
intelligence. In the introduction, we presented two examples of morphological
computation, which described it as the contribution of the body and environment
to a behaviour, that cannot be assigned to any neural system or robot
controller. The
unique information $UI(W':W\setminus A)$ Figure~\ref{fig:mc_2}{\sf B} captures
this notion of morphological computation best, because it vanishes if the
synergistic information $CI(W':W;A)$ (see Fig.~\ref{fig:mc_2}{\sf A}) or the
unique information $UI(W':A\setminus W)$ (see Fig.~\ref{fig:mc_2}{\sf C})
increases. Given Eq.~\eqref{eq:condI-as-sum}, it is clear that the conditional
mutual information $I(W':W|A)$ is positive (see Fig.~\ref{fig:mc_2}{\sf D})
whenever the unique information $UI(W':A\setminus W)$ or the synergistic
information $CI(W':W;A)$ is positive. This is problematic for the following
reason. Figure~\ref{fig:mc_2}{\sf D} show a positive conditional mutual
information $I(W':W|A)$ also for values of $\psi > \phi$, which is
counter-intuitive. Furthermore, as Figure~\ref{fig:mi_vs_ui} shows (note that the
$\phi\psi$ axes are rotated for better visibility), the
conditional mutual information is indifferent for a large range of 
$|\phi - \psi| < d$. Additionally, the conditional mutual information increases
for vanishing $\phi$ and~$\psi$, which again is counter-intuitive, whereas
the unique information $UI(W':W\setminus A)$ (see right-hand side of
Fig.~\ref{fig:mi_vs_ui}) nicely reflects our intuition. Therefore, we conclude
that the unique information $UI(W':W\setminus A)$ is best suited to quantify
morphological computation in the context of embodied artificial intelligence.

% \textbf{Either here or in the discussion we should say: Values of $CI$ are
% larger here; therefore, the difference between $UI$ and the conditional mutual
% information are larger.  Values of $UI$ are better suited to describe MC then
% the conditional mutual information?}

\section{Discussion}
This work proposes a quantification of morphological computation based on an
information decomposition in the sensorimotor loop.
In the introduction, morphological computation was described  as the
contribution of an agent's body and
agent's \emph{Umwelt} to its behaviour. Important to note is that both mentioned examples
highlighted the contribution of the embodiment that resulted solely from
interactions of the body and environment and that cannot be attributed to any
type of control by the agent. This is why we
propose to use a decomposition of the mutual
information $I(W':(W,A))$ into shared, unique and synergistic
information. This allows us to separate contributions of the embodiment from
contributions of the controller (via its actions $A$) and contributions
of both, controller and embodiment.

We showed that the information decomposition is related to our previous
work in the following way. The sum of the unique information
$UI(W':W\setminus A)$ and the synergistic information $CI(W':W;A)$ is
equivalent the conditional mutual information $I(W':W|A)$, which is one of the two earlier
concepts for morphological computation. This relation shows the difference of this work
compared to our former results. We are now able to quantify exactly how much of
the next world state $W'$ is determined by the current world state~$W$, thereby
excluding any influence of the action~$A$. Therefore, we proposed
$UI(W':W\setminus A)$ as a quantification of morphological computation.

In two numerical simulations, we evaluated the decomposition in a parametrised,
binary model of the sensorimotor loop. The world dynamics kernel $\alpha(w'|w,a)$
was parametrised with three parameters, $\phi$, $\psi$, and $\omega$, which
roughly relate to the unique information $UI(W':W\setminus A)$, the 
unique information $UI(W':A\setminus W)$, and the synergistic information
$CI(W':W;A)$. For a fixed value of $\omega$, the two parameters $\phi$ and
$\phi$ were varied to evaluate the information decomposition in the sensorimotor
loop. It was shown that for a vanishing
synergistic parameter $\omega=0$, synergistic
information was present only for $\phi \approx \psi$. This explains
why there is only a marginal difference between $UI(W':W\setminus A)$ and
$I(W':W|A)$ in this setting. For a positive synergistic parameter $\omega=2$, we
saw that the
synergistic information was positive for a much larger domain, which led to a
significant difference between $UI(W':W\setminus A)$ and $I(W':W|A)$. In
particular, the condition mutual information $I(W':W|A)$ was positive
for a larger range of parameter values $\psi$ and~$\phi$. There is a domain
$|\phi - \psi| < d$, for which the conditional mutual information $I(W':W|A)$
is positive and indifferent. One would expect to see a higher morphological
computation mostly when $\phi>\psi$, despite the fact that synergistic information
is present. This shows that the $UI(W':W\setminus A)$ is better suited to
quantify morphological computation.

In~\cite{Zahedi2013aQuantifying} it was proposed that a measure of morphological computation could be used as a guiding
principle in an open-ended self-organised learning setting.  For this to work, this measure should only depend on
information that is intrinsically available to the system.  Clearly, this is not the case for $UI(W':W\setminus A)$.
Therefore, future work will include derivations of the information decomposition, which only include intrinsically
available information.
% , such that the quantification can be used as a first principle objective function in an open-ended
% self-organised learning setting.
It would also be interesting to investigate how
much a formalisation of the information decomposition can benefit from a
consideration of the causal information flow. The starting point for our
decomposition was the mutual information $I(W':(W',A))$, which is a
correlational measure and not a measure of causal dependence, as e.g.~proposed
by \citet{Pearl2000Causality:-Models-Reasoning}.
In currently ongoing work, we are applying the
quantification to motion capturing data of real robots.

\section{Acknowledgements}
This work was partly funded by Priority Program Autonomous Learning (DFG-SPP
1527) of the German Research Foundation (DFG).

\appendix

\section[Computing UI, SI and CI]{Appendix: Computing $UI$, $SI$, and~$CI$.}

In this appendix we shortly explain how we computed the functions $UI$ and~$CI$.
The appendix of~\cite{BROJA13:Quantifying_unique_information} explains how to parametrize the set~$\Delta_{P}$ and how
to solve the optimization problems in the definitions of $UI$, $CI$, and~$SI$.
%  It turns out that this can even be done analytically in the 
In our case, where all variables are binary, $\Delta_{P}$ consists of all probability distributions
$Q_{\gamma_{-1},\gamma_{+1}}$ with
% \begin{equation*}
%   Q_{\gamma_{-1},\gamma_{+1}}(w',w,a) = P(w',w,a) +
%   \begin{cases}
%     \gamma_{w'}, & \text{ if }w=a, \\
%     -\gamma_{w'}, & \text{ if }w\neq a.
%   \end{cases}
% \end{equation*}
\begin{center}
  \begin{tabular}{ccc|l}
    $w'$ & $w$ & $a$ & $Q_{\gamma_{-1},\gamma_{+1}}(w',w,a)$ \\
    \hline
    -1 & -1 & -1 & $P(w',w,a) + \gamma_{-1}\phantom{^{|}}$ \\
    -1 & -1 & +1 & $P(w',w,a) - \gamma_{-1}$ \\
    -1 & +1 & -1 & $P(w',w,a) - \gamma_{-1}$ \\
    -1 & +1 & +1 & $P(w',w,a) + \gamma_{-1}$ \\
    +1 & -1 & -1 & $P(w',w,a) + \gamma_{+1}$ \\
    +1 & -1 & +1 & $P(w',w,a) - \gamma_{+1}$ \\
    +1 & +1 & -1 & $P(w',w,a) - \gamma_{+1}$ \\
    +1 & +1 & +1 & $P(w',w,a) + \gamma_{+1}$ \\
  \end{tabular}
\end{center}
The range of the two parameters $\gamma_{\pm1}$ is restricted in such a way that $Q_{\gamma_{1-},\gamma_{+1}}$ has no
negative entries.  Since every entry $Q_{\gamma_{-1},\gamma_{+1}}(w',w,a)$ involves only one of the two parameters,
$\Delta_{P}$ is a rectangle, bounded by the inequalities
\begin{align*}
  \max\{-P(-1,-1,-1), -P(-1,+1,+1)\} &\le \gamma_{-1}, \\
  \min\{ P(-1,-1,+1), P(-1,+1,-1) \} &\ge \gamma_{-1}, \\
  \max\{-P(+1,-1,-1), -P(+1,+1,+1)\} &\le \gamma_{+1}, \\
  \min\{ P(+1,-1,+1), P(+1,+1,-1) \} &\ge \gamma_{+1}.
\end{align*}

% As shown by~\cite{BROJA13:Quantifying_unique_information}, the optimization problem is convex.
To approximately solve the optimization problem we computed the values on a grid and took the optimal value.  This
simple procedure yields an approximation that is good enough for our purposes.

\end{document}